\documentclass[letterpaper,conference]{IEEEtran}
\IEEEoverridecommandlockouts

\usepackage{graphicx} 
\usepackage{epsfig} 
\usepackage{mathptmx} 
\usepackage{times} 
\usepackage{amsmath} 
\usepackage{amssymb}  
\usepackage{url}

\usepackage{booktabs}
\usepackage{physics}
\usepackage{float}
\usepackage{dblfloatfix}
\usepackage{textcomp}
\usepackage{gensymb}
\usepackage{multicol,tabularx,capt-of}
\usepackage{multirow}

\usepackage{color}

\title{\LARGE \bf
Intrinsic Robotic Introspection: Learning Internal States From Neuron Activations
}

\author{Nikos Pitsillos$^{1}$, Ameya Pore$^{2}$, Bj\o{}rn Sand Jensen$^{1}$ and Gerardo Aragon-Camarasa$^{1}$
\thanks{This research has been supported by EPSRC DTA No. 2279292 and NVIDIA  Corporation  for  the donation of the Titan Xp GPU.}
\thanks{$^{1}$ School of Computing Science, University of Glasgow, UK, {\small n.pitsillos.2@research.gla.ac.uk, bjorn.jensen@glasgow.ac.uk, gerardo.aragoncamarasa@glasgow.ac.uk}}%
\thanks{$^{2}$ Department of Computer Science, University of Verona, Italy, {\small ameya.pore@univr.it}}
}

\begin{document}

\maketitle
\thispagestyle{empty}

\begin{abstract}

We present an introspective framework inspired by the process of how humans perform introspection. Our working assumption is that neural network activations encode information, and building internal states from these activations can improve the performance of an actor-critic model. We perform experiments where we first train a Variational Autoencoder model to reconstruct the activations of a feature extraction network and use the latent space to improve the performance of an actor-critic when deciding which low-level robotic behaviour to execute. We show that internal states reduce the number of episodes needed by about 1300 episodes while training an actor-critic, denoting faster convergence to get a high success value while completing a robotic task.
\end{abstract}

\section{INTRODUCTION\label{sec:intro}}
 
Introspection is a reflective process which allows humans, and potentially artificial systems, to access their internal states. Introspection has been demonstrated to be beneficial in problem-solving \cite{introspection_problem} since it allows us to think or reflect on a given problem and discover new ways to approach it. It draws information from past similar experiences to the task at hand and involves an understanding of our abilities \cite{dehaene}. Prior research \cite{selfawarecomputing} has identified that introspection helps to monitor and optimise a system's behaviour, irrespective of whether the system is biological or in-silico. That is, introspection is the process by which a system observes or accesses its internal states. In this paper, we investigate how we can extract abstract representations of internal states from artificial neural networks and how these representations can be used to inform the execution of a robotic manipulation task. 

Salzman \textit{et al.} \cite{Salzman2010} defined internal states of organisms as a disposition to action, i.e. every aspect of the organism's inner state that contributes to its behaviour. Internal states incorporate environmental stimuli present at a given moment and the temporal context of the stimuli. Each internal state corresponds to one or more states of dynamic variables. Salzman \textit{et al.} \cite{Salzman2010} state that these dynamic variables comprise of neuron activations and synapse weights. The complete set of these dynamic variables constitutes a brain state. We can therefore interpret internal states as higher-level abstractions of the low-level neuron activations. We, therefore, propose to extract internal states by collecting the activations of a neural network to learn an abstract representation that characterises them.

\begin{figure}
    \centering
    \includegraphics[width=0.48\textwidth]{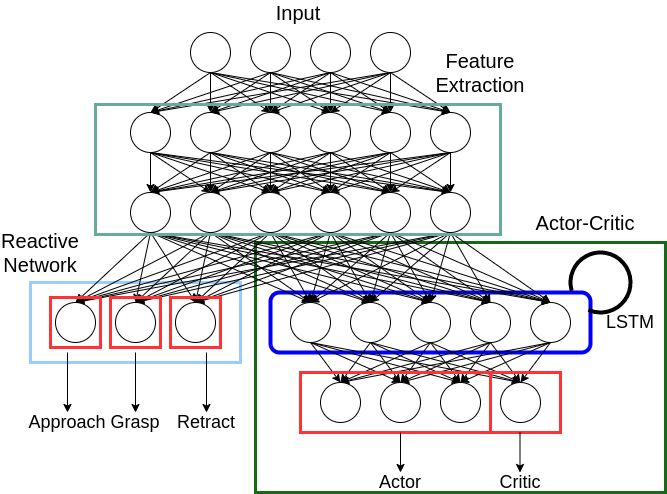}
    \caption{BBRL architecture.  Two fully connected (FC) layers extract features from the environment and a reactive network outputs values for all behaviours given the features. The actor-critic network outputs the value of the state and the action values from which a behaviour is chosen.}
    \label{fig:bbrl}
\end{figure}

\begin{figure*}
    \centering
    \includegraphics[width=0.95\textwidth]{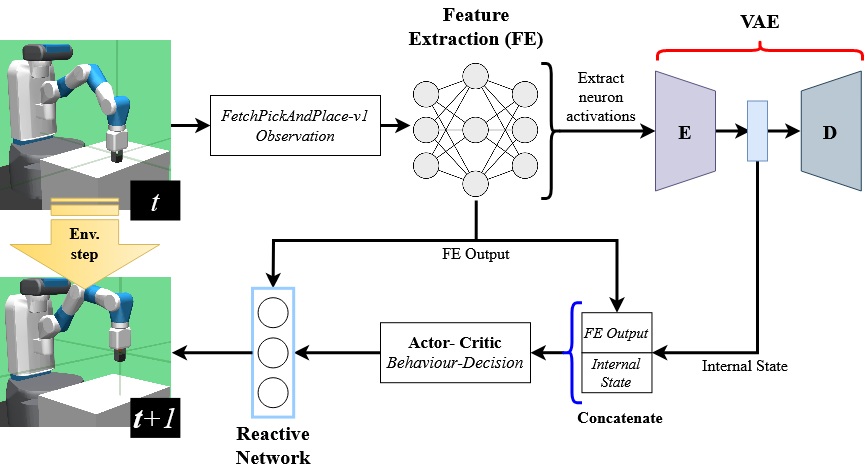}
    \caption{Our approach is inspired by the process of introspection where neuron activations  from a neural network are grouped into internal states.  For this, we use a VAE to learn internal states from a trained neural network.  Internal states are then passed to an actor-critic network to decide what robotic behaviour to execute.}
    \label{fig:approach}
\end{figure*}

In this paper, we investigate how to create artificial internal states and use them for training a deep reinforcement learning (DRL) agent carrying out a manipulation task. That is, the robot perceives its internal states which arise after observing the environment state.  The robot is therefore self-observing its internal states. For this, we use Pore's \textit{et al.} Behaviour Based Reinforcement Learning (BBRL) architecture \cite{bbrl}, Fig. \ref{fig:bbrl}, in which BBRL is used to train a robot to pick up a block in random positions while achieving data-efficiency, i.e. BBRL requires less expert demonstrations to complete a task with respect to the state of the art. Our approach, Fig. \ref{fig:approach}, consists of extracting internal states via a Variational Autoencoder (VAE) to learn a latent space which encodes the neuron activations of the feature extraction component in BBRL.  We then proceed to use the learnt VAE latent space in different ways to investigate how to use the neural network's internal states. We also carry out a series of noise experiments to evaluate the robustness of a learnt internal state during task learning.

Our results show that the use of internal states while training BBRL increases the success of completing the task by $3.6\%$ and reduces training convergence by around $1000$ episodes. We also observe an improvement in the state values of the critic value function. 
The contributions of this paper are thus twofold:

\begin{enumerate}
    \item we propose a framework that extracts and learns internal states of artificial neural networks from low-level neuron activations; and,
    \item demonstrate that these internal states improve the performance of BBRL. 
\end{enumerate}

Our code is available at \url{https://github.com/cvas-ug/bbrl-introspection}. 
The structure of the paper is as follows: Section \ref{sec:background} provides background information of what introspection is motivating the idea behind internal states and discusses other approaches that investigated robotic introspection.  Section \ref{sec:mats} briefly describes the architecture we base our approach on and provides an overview of how we approached the implementation of introspection in a DRL setting.  Section \ref{sec:experiments} describes the experiments we performed and presents our findings.  We finally summarise our work in Section \ref{sec:conclusion} and provide directions for future work.

\section{BACKGROUND}\label{sec:background}

Introspection is closely related to the notion of self, and previous research has demonstrated that the self, and therefore internal states, are shaped by the interactions with the environment \cite{unity}\cite{schillaci}\cite{10.3389/fnbot.2019.00014}. Artificial introspection can be beneficial while training robotic systems to perform a task and allow us to decouple it from the self in order to focus on the advantages of an introspective system. We argue that internal states are intrinsic to the system, and we can build artificial internal states. Hence, we define introspection in this paper as \textit{the process by which a robotic system observes its internal states, from which it gathers information to improve and optimise its behaviour}.


Dehaene \textit{et al.} \cite{dehaene} have stated that an introspective system can monitor its behaviour. Similarly, Agarwal \textit{et al.} \cite{selfawarecomputing} have observed that an introspective system also has the ability to optimise or otherwise improve its behaviour. Given that the introspective process operates on a system's internal states, they are important in the process of monitoring and improving behaviour. Robots are expected to operate in complex environments, and equipping them with the ability to create and observe their internal states can thus enable robots to be flexible and autonomous. That is, robots could have the ability to monitor and observe the correct execution of behaviours and correct failures during execution. For instance, Popov \textit{et al.} \cite{popov} explored robotic introspection by using a genetic algorithm to evolve internal states. Their work used a similar definition to ours since they built internal states from a robot control system, computing resources, and their changing states. Our focus is to create internal states from the features learnt by a neural network and create high-level representations of the network's activations.

Similarly, Infantino \textit{et al.} \cite{infantino} implemented an introspective architecture to enable a robot to communicate its internal states verbally. In their work, internal states are created from information coming from software modules and their corresponding software documentation, running processes in the robotic system, and hardware specification. Infantino's \textit{et al.} work builds introspection into a robot by enabling it to examine its running processes. However, we argue that internal states should be constructed by taking into account the robot's interactions with the environment at the same time as monitoring its intrinsic operation.

Rojas \textit{et al.} \cite{Rojas_2017} explored robotic introspection by extending the \textit{`sense-plan-act'} paradigm to \textit{`sense-plan-act-verify'}. This paradigm allows the robot to understand its high-level behaviour from multi-modal signals to classify nominal and abnormal states of the robot and recover from them. Similarly, Wu \textit{et al.} \cite{Wu_2017}\cite{Wu2017LearningRI} investigated the modelling ability of Bayesian non-parametric techniques on Markov Switching Processes to provide the robot with temporal introspection about its evolving state \cite{Wu_2017}. The latter enabled the robot to make timely decisions to ensure a goal is achieved even in the presence of unexpected events. However, we address the introspection problem from an intrinsic point of view, since we build internal states intrinsically from the robot's understanding of the environment and use them for training a robot to pick and place a block. 
Introspection has not been explored in a deep reinforcement learning framework since the above approaches \cite{Rojas_2017}\cite{Wu_2017}\cite{Wu2017LearningRI} have implemented introspection as software modules that evaluate the system's internal execution, and represent an outside view of introspection as defined in \cite{dehaene}.  We, therefore, devise a framework of robotic introspection where we focus on how internal states can be created intrinsically from the robot's state.  Previous approaches treat the introspection problem as an observer where the robot's state is to be described during the  execution of the robot. In our work we take an intrinsic view of introspection where internal states are constructed by observing the environment state and inform the agent's behaviour. Finally, our approach uses information from the model's execution rather than the output of the model.


\section{Introspection in BBRL}\label{sec:mats}


Our hypothesis is that \textit{activations in a neural network encode internal states that can be utilised to inform the robotic task being executed}. The goal of this paper is to identify structure in the activations of a neural network, create abstract representations of these activations and use them to improve the robot's behaviour.

As noted in Section \ref{sec:intro}, we use the BBRL architecture \cite{bbrl} to test our hypothesis. BBRL is composed of three main parts: (1) a feature extraction network (FE) with two FC layers, (2) a reactive network which outputs action movements in 3D space for approach and retract behaviours as well as a rotation action for rotating the end-effector while grasping, and (3) an actor-critic architecture that learns to choreograph these behaviours temporally. Fig. \ref{fig:bbrl} shows a diagram of the architecture. The reactive and actor-critic networks share the same FE network.


At the lowest level, introspection occurs in neuron activations which can be grouped into internal states \cite{Salzman2010} denoting a disposition to action. Hence, we propose to use a Variational Autoencoder (VAE) \cite{vae} to learn low-dimensional representations of a neural network's internal states. This low-dimensional representation describes the internal state of the robot. Given that internal states are created using a subset of the activations and neuron synapse weights, a VAE allows us to identify the features that describe the low-level neural network state and, consequently, equip BBRL with introspection capabilities. Therefore, our approach, as shown in Fig. \ref{fig:approach}, consists of training the VAE with activations from the FE. While training BBRL, we used the VAE's encoder to create an internal state which is then concatenated to the features from the FE and passed to the actor-critic to decide which behaviour to execute, given a state. BBRL is trained as reported in \cite{bbrl}. We summarise BBRL's training strategy below for completeness. 

The training strategy for BBRL consists of training the FE and approach behaviour first while using expert demonstrations for grasp and retract. We then train the grasp behaviour with frozen weights for the FE and using expert demonstrations for approach and retract and, finally, repeat for retract. BBRL's actor-critic network is thus trained using pretrained FE and reactive networks.

In this paper, we focus on the performance of the actor-critic. Therefore, we use a pretrained FE and reactive network weights for training our VAE. For this, we collect a dataset of activations from the FE by running the BBRL's feature extraction and reactive networks using expert demonstrations for $2000$ episodes. After creating the activations dataset, we train the VAE for $100$ epochs with an Adam optimiser. We use an $80$/$20$ split of the dataset. The input to the VAE is of size $256$, which is the total size of the two FC layers in the FE. The encoder consists of four FC layers. The first two layers in the encoder have $400$ and $128$ features, respectively, and the output is passed through an ELU activation function. The final two layers output the latent mean and log variance with a hidden latent size of $50$. The decoder consists of three FC layers with $128$, $400$ and $256$ features respectively, that decodes a random sample drawn from the latent code distribution. We use the following loss function to train the VAE in all experiments, Section \ref{sec:experiments}.




\begin{equation}
    L_{VAE} = MSE(x_{real}, x_{recons}) + D_{KL}(q(z|x_{real})||p(z))
    \label{eq:vae}
\end{equation}

where $MSE(x_{real}, x_{recons})$ is the mean-squared error between the input activations from the FE, $x_{real}$, and their reconstruction, $x_{recons}$. $D_{KL}(q(z|x_{real})||p(z))$ is the Kullback-Leibler divergence between the distribution of the VAE's encoder network and the true distribution of the latent variables, $z$.


\section{EXPERIMENTS\label{sec:experiments}}


In this work, we create internal states that are intrinsic to BBRL. That is, internal states are created from within the network's hidden neuron activations which result from the robot's interactions with the environment. Experiment 1, \ref{ssec:exp_1} uses a VAE to learn the internal states of BBRL by reconstructing the FE's neuron activations to understand whether the internal states, i.e. the VAE latent space, have a structure that represents robotic behaviours. We thus use t-SNE \cite{Maaten08visualizingdata} to visualise the learnt latent space. Salzman \textit{et al.} \cite{Salzman2010} state that internal states play an important role in decision making. Experiment 2, \ref{ssec:exp_2}, thus explores how the learnt internal states can be used to improve the performance of BBRL. This experiment evaluates the representational richness of the learnt latent space. Finally, Experiment 3, \ref{ssec:exp_3}, simulates a real-world scenario by adding noise to the input of BBRL and explores the effect on the performance of the actor-critic when internal states are learnt in the presence of noise.

\begin{figure}
    \centering
    \includegraphics[scale=0.35]{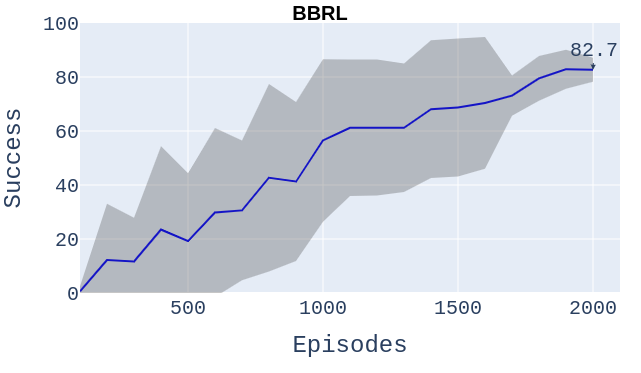}
    \caption{Original BBRL actor-critic success (in percentages) for 2000 episodes. Training follows same procedure as in \cite{bbrl} using a pretrained FE network.}
    \label{fig:orig}
\end{figure}

BBRL was demonstrated to reach 100\% success on the pick and place task \cite{bbrl}. In this paper, we, therefore, focus on the first 2000 episodes since the focus of this paper is to investigate how internal states can be used to train the actor-critic. Training beyond 2000 episodes for all models used in this paper can be found at \url{https://github.com/cvas-ug/bbrl-introspection}. Fig. \ref{fig:orig} shows the training performance of training BBRL's actor-critic for the first 2000 episodes.

All experiments where performed in the FetchPickAndPlace-v1 OpenAI gym environment \cite{gym} using MuJoCo 1.5 \cite{mujoco}. The experiments were run on a desktop PC with an AMD Ryzen 9 3900X CPU, an RTX 2080 Super GPU and 32 GB RAM running Ubuntu 20.04.

\subsection{Experiment 1: Learning Internal States}\label{ssec:exp_1}

\begin{figure}
    \centering
    \includegraphics[width=0.48\textwidth]{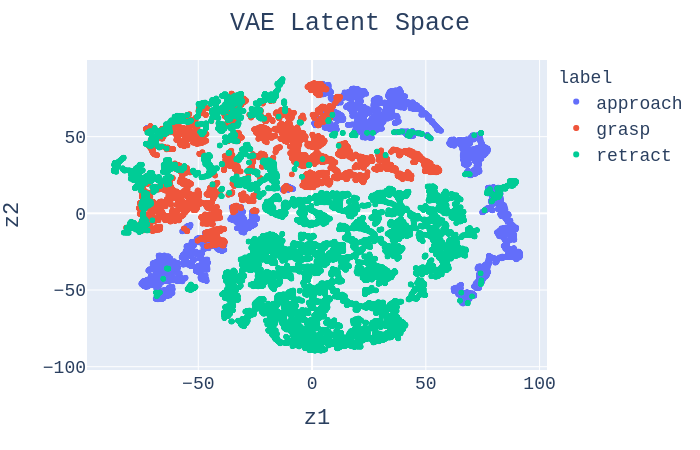}
    \caption{Two component t-SNE visualisation of the learnt latent space using a two component t-SNE.  $z1$ and $z2$ correspond to the two components returned by t-SNE describing the data.}
    \label{fig:vae-latent}
\end{figure}

To build an internal state for BBRL, we execute the reactive network using hardcoded behaviours.  At each timestep we collect the resulting hidden activations when the environment observation is feed-forwarded through the reactive network. Each vector of hidden activations is stored along with its corresponding behaviour to create an activations dataset.  We then train a VAE to reconstruct the hidden activations following the strategy discussed in Section \ref{sec:mats}.  t-SNE \cite{Maaten08visualizingdata} was used to visualise the latent space, and the behaviour labels from the activations dataset were used to label the latent space points.  Fig. \ref{fig:vae-latent} shows the latent space of the VAE when encoding the validation dataset.

From Fig. \ref{fig:vae-latent}, we can observe that there is not a clear grouping in the activations. However, one interesting aspect is how the latent codes for the behaviours appear to merge between each other, denoting transitions between the robot's behaviours. That is, the latent codes for approach (blue) appear to merge into the latent codes for grasp (red) and likewise for grasp and retract (green). This indicates that the latent space has encoded a nonlinear representation of the internal states, and organises the temporal execution of the behaviours into a static representation that can be employed during task execution.

\begin{figure*}
    \centering
    \includegraphics[width=0.81\textwidth]{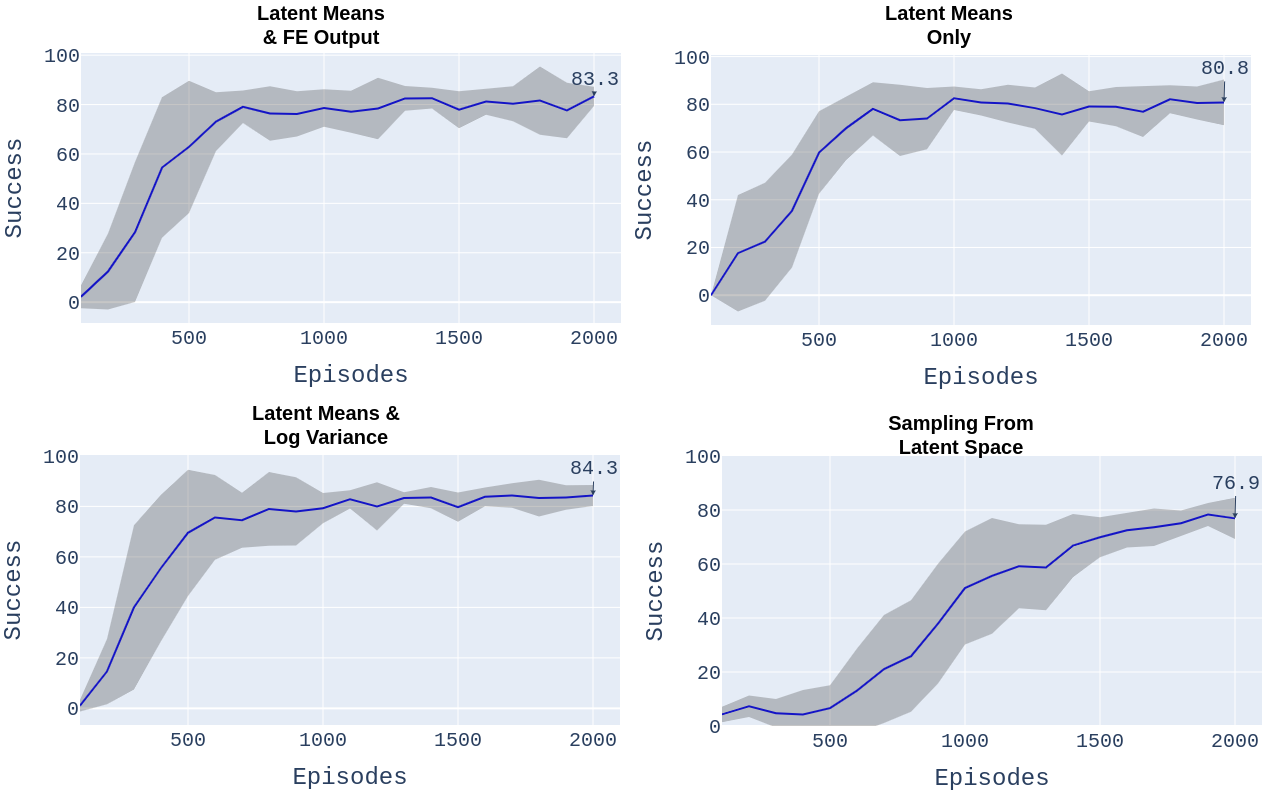}
  \caption{Average success in percentage over 10 runs for all proposed experiments.}
  \label{fig:avg_success} 
\end{figure*}

\subsection{Experiment 2: Investigating Internal State Effectiveness}\label{ssec:exp_2}

As stated in Section \ref{sec:experiments}, we propose four variants of using internal states to train the actor-critic.  Through these experiments, we aim to identify which method is more effective in enabling to train BBRL's actor-critic. In the following experiments, the VAE encoder layers were fine-tuned while training BBRL's actor-critic.

\subsubsection{Concatenating Latent Means with Feature Network Outputs}

This experiment aims to answer the question of \textit{whether internal states can act as supplementary information to the actor-critic}. During training, we collect the FE's activations and encode them to an internal state using the trained VAE. The VAE latent means are then concatenated to the FE second layer output, and the resulting vector is used to train the actor-critic.  This variant results in a similar performance to BBRL in terms of success. However, Fig. \ref{fig:avg_success} shows that training converges faster (i.e. reaches to $80\%$ in around 700 episodes) whereas the original BBRL reaches $80\%$ after 2000 episodes. Furthermore, the variance in the success value is less when comparing with the original BBRL in Fig. \ref{fig:orig}.

\subsubsection{Using Latent Means Only}

This experiment aims to answer the question of \textit{whether internal states are a rich enough representation of the FE state to be used on their own to train the actor-critic}. At training, we collect the FE's activations and encode them to an internal state. The VAE latent means are then passed to the actor-critic as input. The actor-critic network receives no other information as input. Fig. \ref{fig:avg_success} shows a slightly less success value than the original, Fig. \ref{fig:orig}. However, the number of episodes to convergence, as well as the variance in success is lower throughout training except during the final stages. Similar to the experiment in \ref{ssec:exp_1}, this approach converges to approximately $80\%$ at around 700 episodes. This supports our hypothesis that the actor-critic can be trained only by using internal states since it has managed to converge to a high enough success value comparable to the original BBRL, Fig. \ref{fig:orig}.

\subsubsection{Concatenating Latent Means \& Log Variance}

This experiment followed a similar approach to the above experiment and aims to answer the same question using a different input to the actor-critic. Here, we concatenate the output of the encoder means and log variances into a single vector which is passed to the actor-critic. As shown in Fig. \ref{fig:avg_success}, this approach is the fastest to converge at around $600$ episodes and also results in a slightly higher success value during training. It also reduces the variance to a great extent, further supporting our hypothesis that internal states are sufficient to train the actor-critic.

\subsubsection{Sampling From Latent Space}

The question this experiment attempts to answer is \textit{whether it is possible to train the actor-critic by sampling from the latent space}. In this experiment, the input to the actor-critic is obtained by first encoding the FE's activations and then sampling an internal state from the VAE latent space. Fig. \ref{fig:avg_success} shows the result after training the actor-critic.  This approach takes about the same number of episodes to converge and results in a slightly lower success value than the original BBRL, Fig. \ref{fig:orig}.  This can be explained by the fact that sampling from the latent space introduces a random component when sampling from the latent space distribution.

\begin{figure*} 
    \centering
  \includegraphics[width=0.95\textwidth]{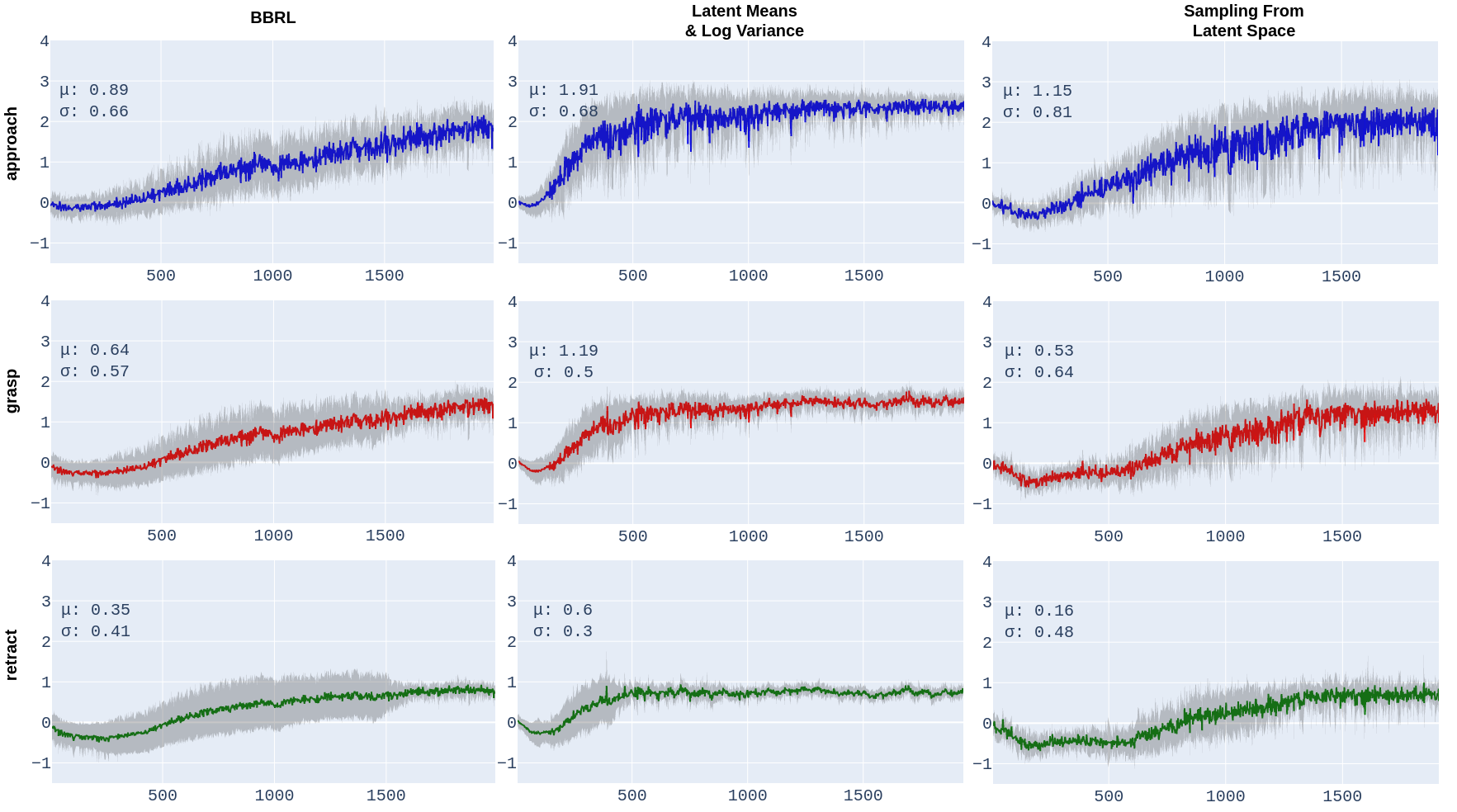}
  \caption{State values obtained from the critic output.
  Training BBRL with internal states results in faster state value convergence and improves the state value for all behaviours over the original BBRL.}
  \label{fig:state_values}
\end{figure*}

\begin{figure*}
    \centering
    \includegraphics[width=0.95\textwidth]{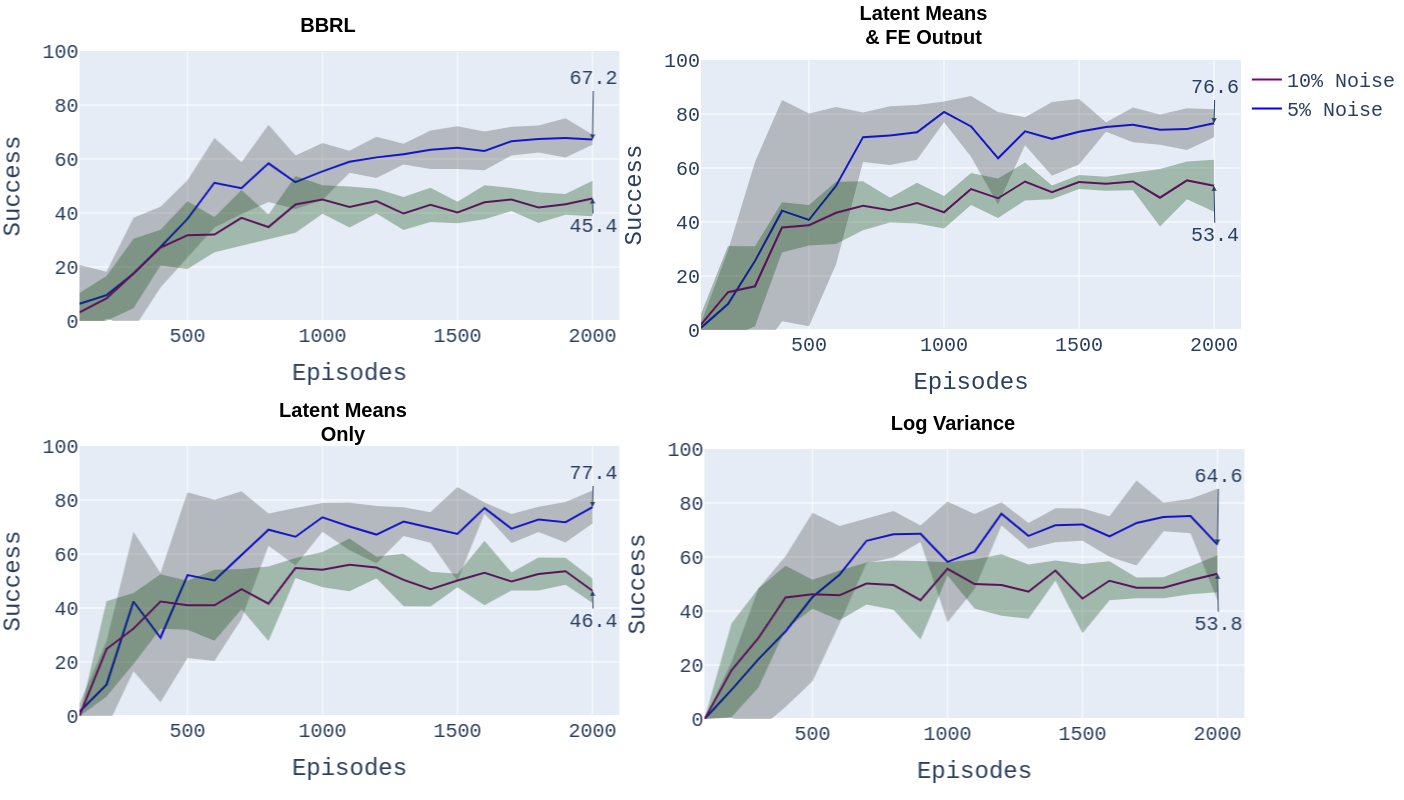}
  \caption{Average success in percentage over 5 runs with two noise levels. }
  \label{fig:avg_success_noise} 
\end{figure*}

\subsection{State Value Results}
To perform a more comprehensive analysis, we plot the state values over 2000 episodes of training.  In the robotic task used in this paper and, consequently, BBRL, there are three possible states for the robot, which are: approach, grasp, and retract. In the actor-critic architecture, the critic implements a value function which is the total amount of reward an agent is expected to accumulate from that state onward \cite{sutton}.  If our hypothesis holds (i.e. training with intrinsic internal states improves BBRL's performance), then we should also expect to see an improved state value over behaviours.

Fig. \ref{fig:state_values} shows how the state values change during training when no noise affects the input. As observed in Fig. \ref{fig:state_values}, the experiment where we concatenate the latent means and log variances and use them as input to the actor-critic results in a higher state value for all behaviours whereas sampling from the latent space results in a low state value. It is worth noting that training BBRL with internal states results in faster convergence for the state values at around 500 episodes of training.

\subsection{Experiment 3: Investigating Robustness to Noise}\label{ssec:exp_3}

This experiment aims to replicate the noise when sensing the state of the robot and environment in a real setting, and answer the question of \textit{which model variant is most robust to noise}. For this, we first train BBRL for 2000 episodes where the input is affected by noise drawn from $\mathcal{N}$(0,0.1). The noise sample drawn is of the same size as the environment observation, and we train with two noise levels, 5\% and 10\%. To ensure we learnt the correct internal state representations, we then run the trained BBRL model using expert demonstrations resulting from training with the two noise levels to collect datasets of activations corresponding to both noise levels. We then proceed to train two VAE models, one for each noise level.

We perform experiments with noise on all experiment variants, including the original BBRL. Fig. \ref{fig:avg_success_noise} shows the results of this experiment.  We omit the experiment where we sample an internal state from the latent space because it did not show any evidence of success with a noise level of 5\%. All proposed approaches result in a higher success value for both noise levels except when latent means and log variances are concatenated.  However, all approaches converge faster than BBRL and have a larger success value throughout training. These results provide evidence of the robustness of using internal states to train the actor-critic.



\section{CONCLUSIONS}\label{sec:conclusion}

In this paper, we have trained an actor-critic network that learns to choreograph low-level behaviours via internal states of an artificial neural network. We also investigated how we can learn abstract representations of low-level activations from a neural network and then use them to improve training performance. In our experiments, we found that by using internal states as the input to the actor-critic, we obtained a reduction of $1300$ episodes while training the actor-critic as in \cite{bbrl}.
To accomplish this, we have used a VAE to learn internal states from low-level activations and proposed four variants to use them. Our results showed that the combination of latent space means and log variance as the input to an actor-critic network improves the number of episodes to convergence, reduces the variance of success during training, and results in a larger success value. We also showed further evidence of the benefits when training the actor-critic with internal states as there is an increased state value for all behaviours by $1.02$, $0.55$ and $0.25$ for approach, grasp and retract, Fig. \ref{fig:state_values}, respectively, with respect to \cite{bbrl}.

Internal states therefore improve the performance of BBRL since they incorporate the environment observation and the hidden activations of the feature extraction network which form part of the dynamic variables discussed in Section \ref{sec:intro}.  These dynamic variables seem to be more informative than only propagating the input in the network enabling faster learning of the task. This is supported by Cleeremans \textit{et al.} \cite{CLEEREMANS20071032} where a neural network trained to reconstruct another neural network's activations begins to learn faster than the network it simulates.  Finally, we performed a statistical significance analysis to compare BBRL with the two best performing Experiment 2 variants. Following \cite{DBLP:journals/corr/abs-1806-08295}, we used Welch's test to compare the average performance from 5 random seeds.  Experiment 2A and 2C both pass Welch's test with a p-value of 0.0259 and 0.0119 respectively.

Future work consists of exploring whether we can also use internal states to dictate the behaviour of BBRL. Instead of using an actor-critic, we aim to adopt an unsupervised learning approach to train our robot. 
We further aim to explore how internal states can be evolved using evolutionary techniques \cite{popov} to both improve performance but also acquire new skills. 

\section*{ACKNOWLEDGEMENT}
We thank Paul Siebert, Ali Al-Qallaf, Piotr Ozimek, Li Duan, Ozan Bahadir, Vanja Popovic and Lewis Boyd for valuable discussions at the earlier stages of this research.

\bibliographystyle{IEEEtran}
\bibliography{References}
\end{document}